\documentclass[10pt,twocolumn,letterpaper]{article}

\usepackage{iccv}
\usepackage{times}
\usepackage{epsfig}
\usepackage{graphicx}
\usepackage{amsmath}
\usepackage{amssymb}
\usepackage{subfigure}
\usepackage{multirow}

\usepackage{soul}

\usepackage[pagebackref=true,breaklinks=true,letterpaper=true,colorlinks,bookmarks=false]{hyperref}

\iccvfinalcopy 

\def\iccvPaperID{1809} 

\ificcvfinal\fi

\begin{document}

\title{Multimodal Convolutional Neural Networks for Matching Image and Sentence}

\author{Lin Ma\qquad Zhengdong Lu\qquad Lifeng Shang\qquad Hang Li\\
Noah's Ark Lab, Huawei Technologies\\
{\tt\footnotesize forest.linma@gmail.com, \{Lu.Zhengdong, Shang.Lifeng, HangLi.HL\}@huawei.com }
}

\maketitle

\begin{abstract}
  In this paper, we propose multimodal convolutional neural networks ($m$-CNNs) for matching image and sentence.  Our $m$-CNN provides an end-to-end framework with convolutional architectures to exploit image representation, word composition, and the matching relations between the two modalities. More specifically, it consists of one image CNN encoding the image content, and one matching CNN learning the joint representation of image and sentence. The matching CNN composes words to different semantic fragments  and learns the inter-modal relations between image and the composed fragments at different levels, thus fully exploit the matching relations between image and sentence. Experimental results on benchmark databases of bidirectional image and sentence retrieval demonstrate that the proposed $m$-CNNs can effectively capture the information necessary for image and sentence matching.  Specifically, our proposed $m$-CNNs for bidirectional image and sentence retrieval on Flickr30K and Microsoft COCO databases achieve the state-of-the-art performances.


\end{abstract}


\section{Introduction}
Associating image with natural language sentence plays the essential role in many applications. Describing the image with natural sentences is useful for image annotation and caption \cite{gong_eccv2014,kiros_nips2012,ordonez_nips2011}, while retrieval image with query sentences is more convenient and helpful for the natural image search applications \cite{hodosh_jair2013,karpathy_2014}. The association between image and sentence can be formalized as a multimodal matching problem, where the semantically correlated image and sentence pairs should produce higher matching scores than uncorrelated ones. 

\begin{figure}[t]
\begin{center}
\includegraphics[width=1\linewidth]{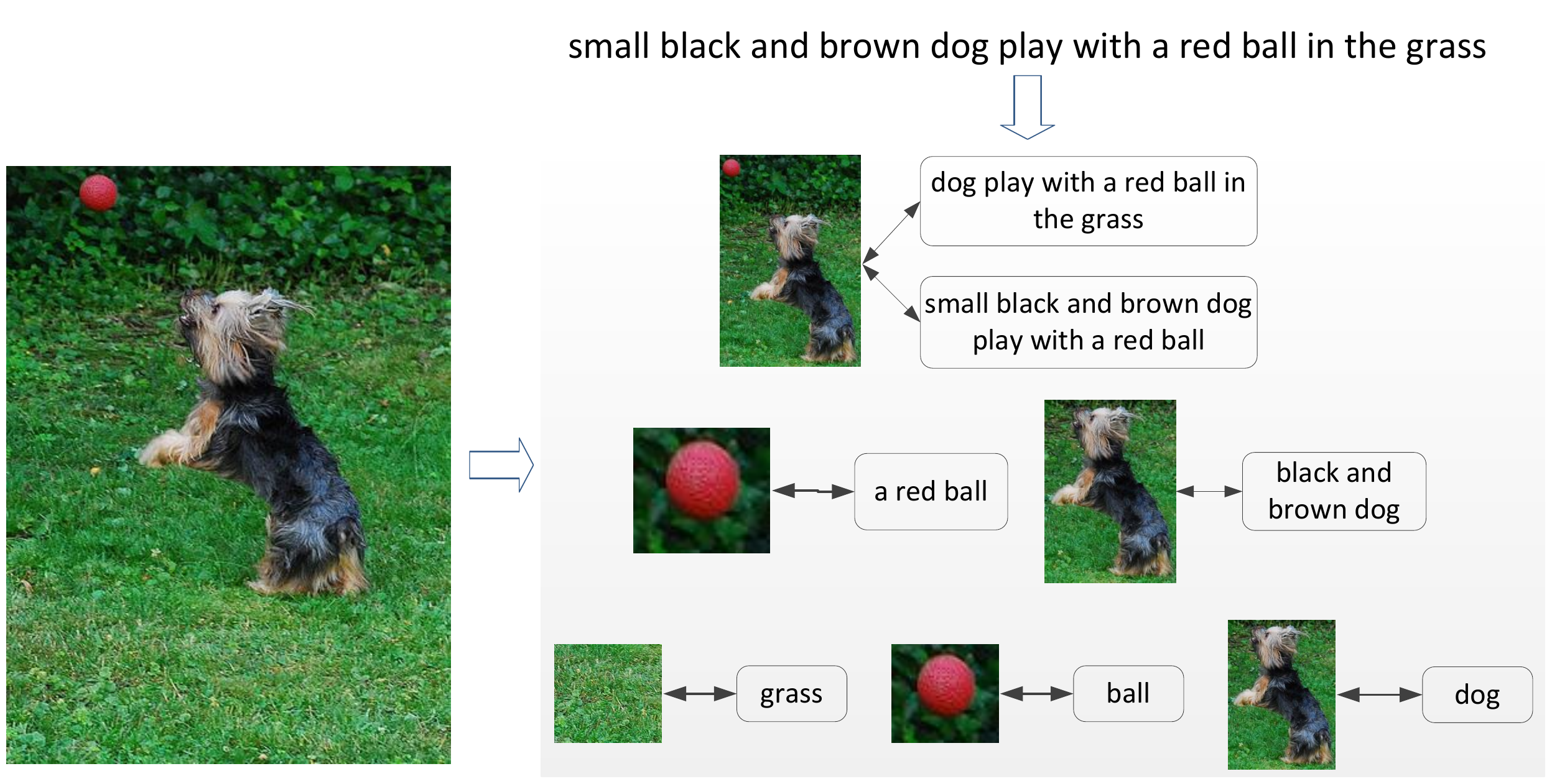}

\end{center}
   \caption{The multimodal matching relations between image and sentence. The words and phrases, such as ``\texttt{\small grass}", ``\texttt{\small a red ball}", and ``\texttt{\small small black and brown dog play with a red ball}", correspond to the image areas of their grounding meanings. The global sentence ``\texttt{\small small black and brown dog play with a red ball in the grass}" expresses the whole semantic meaning of the image content.}
\label{fig1}
\end{figure}


The multimodal matching relations between image and sentence are complicated, which happen at different levels as shown in Figure \ref{fig1}. The words in the sentence, such as ``\texttt{\small grass}", ``\texttt{\small dog}", and ``\texttt{\small ball}", denote the objects in the image. The phrases describing the objects and their attributes or activities, such as ``\texttt{\small black and brown dog}", and ``\texttt{\small small black and brown dog play with a red ball}", correspond to the image areas of their grounding meanings. The whole sentence ``\texttt{\small small black and brown dog play with a red ball in the grass}", expressing a complete semantic meaning, associates with the whole image content.  These matching relations should be all taken into consideration for an accurate inter-modal matching between image and sentence. Recently, much research work focuses on  modeling the image and sentence matching relation at the specific level, namely the word level \cite{nitish_icml2012,nitish_nips2012,frome_nips2013}, phrase level \cite{zitnick_iccv2013,sadeghi_cvpr2011}, and sentence level \cite{hodosh_jair2013, karpathy_2014, socher_tacl2014}. However, to the best of our knowledge, there are no models to fully exploit the matching relations between image and sentence by considering the word, phrase, and sentence level inter-modal correspondences together.

The multimodal matching between image and sentence requires good representations of the image and sentence. Recently, deep neural networks have been employed to learn better image and sentence representations. Specifically, convolutional neural networks (CNNs) have shown their powerful abilities on image representation \cite{he_eccv2014,simonyan_arxiv2014,szegedy_arxiv2014,he_arxiv2015} and sentence representation \cite{kalchbrenner_acl2014,ykim_emnlp2014}. However, the ability of CNN on multimodal matching, specifically the image and sentence matching problem, has not been studied.

In this paper, we propose a novel multimodal convolutional neural network ($m$-CNN) framework for the image and sentence matching problem. By training on a set of image and sentence pairs, the proposed $m$-CNNs are able to retrieve and rank the images given a natural sentence query, and vice versa. Our core contributions are:
\begin{enumerate}
  \item CNN is firstly studied for the image and sentence matching problem. We employ convolutional architectures to summarize the image, compose words of the sentence into different semantic fragments, and learn the matching relations and interactions between image and the composed fragments.
  \item The complicated matching relations between image and sentence are fully studied in our proposed $m$-CNN by letting image and the composed fragments of the sentence meet and interact at different levels. We validate the effectiveness of $m$-CNNs on bidirectional image and sentence retrieval experiments, in which we achieve performances superior to the state-of-the-art approaches.
\end{enumerate}

\section{Related Work}
\label{sec_related}
\subsection{Association between Image and Text}
There is a long thread of work on the association between image and text. Early work usually focuses on modeling the correlation between image and the annotating words \cite{frome_nips2013, nitish_icml2012, nitish_nips2012,grangier_icann2006, weston_ijcai2011} or phrases \cite{sadeghi_cvpr2011, zitnick_iccv2013}. These models cannot well capture the complicated matching relations between image and the natural sentence. Recently, the association between image and sentence has been studied for bidirectional image and sentence retrieval \cite{hodosh_jair2013, karpathy_2014, socher_tacl2014, yan_cvpr2015, klein_cvpr2015, kiros_2015, plummer_2015} and automatic image captioning \cite{chen_2014,donahue_2014,karpathy_dvsa_2014,kiros_icml2014,kiros_2014,mao_2014,mao_iclr_2015,vinyals_2014}.

For bidirectional image and sentence retrieval, Hodosh \textit{et al.} \cite{hodosh_jair2013}  proposed KCCA to discover the shared feature space between image and sentence. However, the highly non-linear inter-modal relations cannot be well exploited based on the shallow representations of image and sentence.
Recent papers seek better representations of image and sentence from deep architectures. Socher \textit{et al.}  \cite{socher_tacl2014} proposed to employ the semantic dependency-tree recursive neural network (SDT-RNN) to map the sentence into the same semantic space as the image representation, and the association is then measured as the distance in that space. Yan \textit{et al.} \cite{yan_cvpr2015} stacked fully connected layers together to represent the sentence and used deep canonical correlation analysis (DCCA) for matching images and text. Klein \textit{et al.} \cite{klein_cvpr2015} used the Fisher vector (FV) for the sentence representation. Kiros \textit{et. al} \cite{kiros_2015} proposed skip-thought vector (STV) to encode the sentence for matching the image. As such, the global level matching relations between image and sentence are studied by representing the sentence as a global vector. However, they neglect the local fragments of the sentence and their correspondences to the image content. Compared with \cite{socher_tacl2014}, Karpathy \textit{et al.} \cite{karpathy_2014} work on a finer level by aligning the fragments of sentence and regions of image. Plummer $et. al$ \cite{plummer_2015} used the entities to collect region-to-phrase (RTP) correspondences for richer image-to-sentence models. The local inter-modal correspondences between image and sentence fragments are thus studied, where the global matching relations are not considered. As illustrated in Figure \ref{fig1}, the image content corresponds to different fragments of sentence from local words to the global sentence. To fully exploit the inter-modal matching relations,  we propose $m$-CNNs to compose words of sentence to different fragments, let the fragments meet image at different levels, and learn their matching relations.

For automatic image captioning,  the authors use recurrent visual representation (RVP) \cite{chen_2014}, multimodal recurrent neural network ($m$-RNN) \cite{mao_2014,mao_iclr_2015}, multimodal neural language model (MNLM) \cite{kiros_icml2014,kiros_2014}, neural image caption (NIC) \cite{vinyals_2014}, deep visual-semantic alignments (DVSA) \cite{karpathy_dvsa_2014}, and long-term recurrent convolution networks (LRCN) \cite{donahue_2014} to learn the relation between image and sentence and generate the caption for a given image. Please note that those models naturally produce scores for image-sentence association (e.g., the likelihood of a sentence as the caption for a given image). It can thus be readily used for bidirectional retrieval.

\subsection{Image and Sentence Representation}
For image, CNNs have demonstrated their powerful abilities to learn the image representation from image pixels, which achieved the state-of-the-art performances on image classification \cite{he_eccv2014,simonyan_arxiv2014,szegedy_arxiv2014,he_arxiv2015} and object detection \cite{ouyang_arxiv14, girshick_cvpr2014}. For sentence, there is a thread of neural networks for the sentence representation, such as CNN \cite{kalchbrenner_acl2014,ykim_emnlp2014}, time-delay neural network \cite{collobert_jmlr2011}, recursive neural network \cite{irsoy_nips2014}, and recurrent neural network \cite{blunsom_arxiv2013,socher_tacl2014,mao_2014,sutskever_icml2011}. The obtained sentence representation can be used for the sentence classification \cite{ykim_emnlp2014}, image and sentence retrieval \cite{socher_tacl2014,mao_2014}, language modeling \cite{collobert_jmlr2011}, text generation \cite{karpathy_dvsa_2014,sutskever_icml2011}, and so on.

\section{${m}$-CNNs for Matching Image and Sentence}
\label{sec_model}

\begin{figure}[t!]
\begin{center}
   \includegraphics[width=1\linewidth]{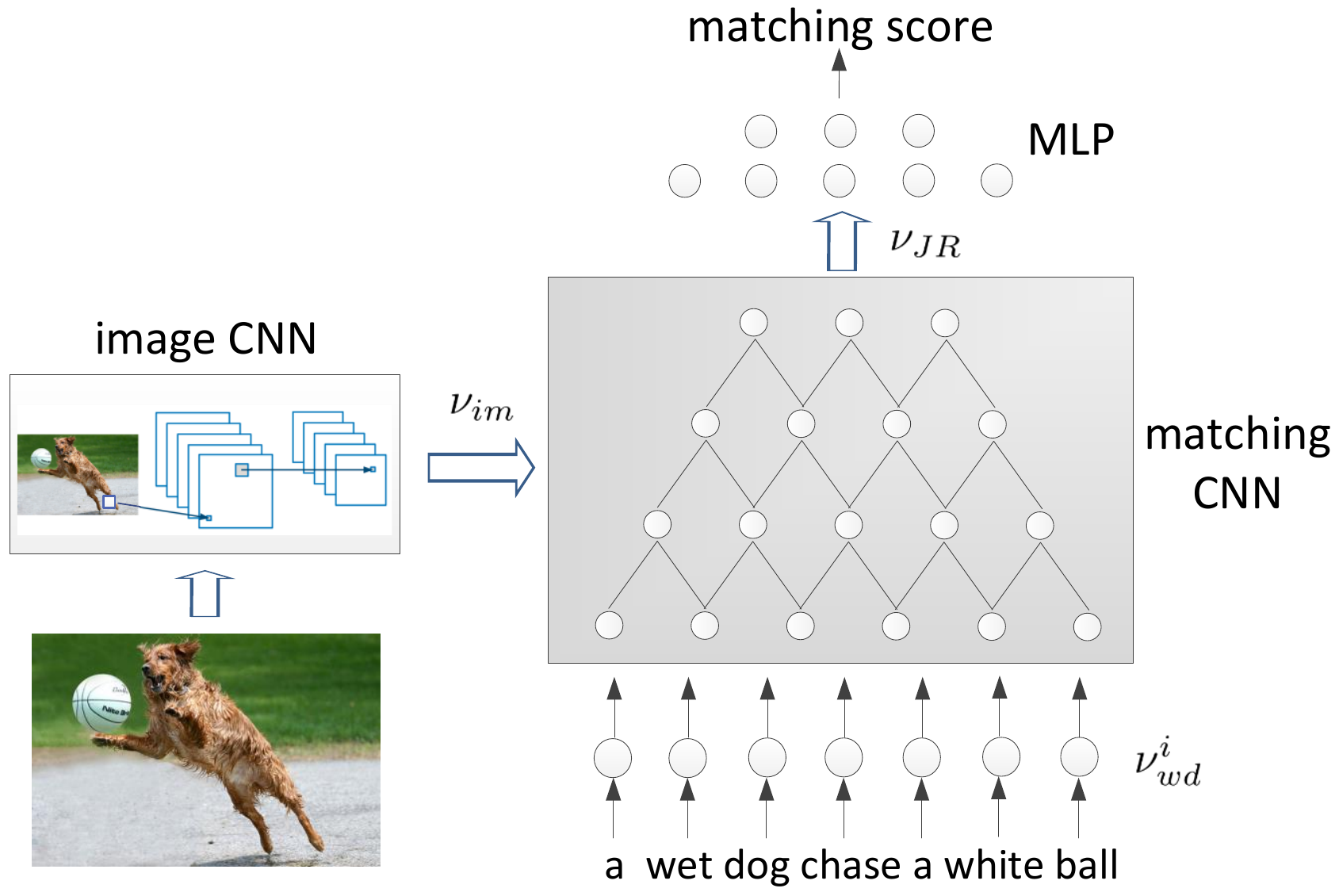}

\end{center}
   \caption{The $m$-CNN architecture for matching image and sentence. Image representation is generated by the image CNN. Matching CNN composes words to different fragments of the sentence and learns the joint representation of image and sentence fragments. MLP summarizes the joint representation and outputs the matching score.}
\label{fig:framework}
\end{figure}

As illustrated in Figure \ref{fig:framework}, $m$-CNN takes the image and sentence as the inputs and generates the matching score between them. More specifically, $m$-CNN consists of three components.
\vspace{-5pt}
\begin{itemize}
  \item {\bf Image CNN:} The image CNN is used to generate the image representation for matching the fragments composed from words, which is computed as follows:
\begin{equation}
\label{eq_image_cnn}
\begin{split}
\nu_{im} = \sigma(\mathbf{w}_{im}(CNN_{im}(I))+b_{im}),
\end{split}
\end{equation}
where $\sigma(\cdot)$ is the activation function (e.g., Sigmoid or ReLU~\cite{relu}). $CNN_{im}$ is an image CNN which takes the image as the input and generates a fixed length image representation. The successful image CNNs for image recognition, such as \cite{sermanet_arxiv2014,simonyan_arxiv2014}, can be used to initialize the image CNN, which returns the 4096-dimensional activations of the fully connected layer immediately before the last ReLU layer. The matrix $\mathbf{w}_{im}$ is of the dimension $d\times4096$, where $d$ is set as $256$ in our experiments. Each image is thus represented as one $d$-dimension vector $\nu_{im}$.
\item {\bf Matching CNN}  The matching CNN takes the encoded image representation $\nu_{im}$ and word representations $\nu_{wd}^i$ as the input and produces the joint representation $\nu_{JR}$. As illustrated in Figure \ref{fig1}, the image content may correspond to sentence fragments with varying scales, which will be adequately considered in the learnt joint representation of image and sentence. Targeting at fully exploiting the inter-modal matching relation, our proposed matching CNNs firstly compose words to different semantic  fragments and then let the image meet these fragments to learn their inter-modal structures and interactions. More specifically, different matching CNNs are designed to make the image interact with the composed fragments at different levels to generate the joint representation, from the word and phrase level to the sentence level. Detailed information of the matching CNNs at different levels will be introduced in the following subsections.
\item {\bf MLP} Multilayer perceptron (MLP) takes the joint representation $\nu_{JR}$ as the input and produces the final matching score between image and sentence, which is calculated as follows.
\begin{equation}
\label{eq_mlp}
\begin{split}
\mathbf{s}_{match} = \mathbf{w}_{s}\big( \sigma(\mathbf{w}_{h}(\nu_{JR})+b_{h})\big )+b_{s}.
\end{split}
\end{equation}
where $\sigma(\cdot)$ is the nonlinear activation function. $\mathbf{w}_{h}$ and $b_{h}$ are used to map $\nu_{JR}$ to the representation in the hidden layer. $\mathbf{w}_{s}$ and $b_{s}$ are used to compute the matching score between image and sentence.

\end{itemize}

The three components of our proposed $m$-CNN are fully coupled in the end-to-end image and sentence matching framework, with all the parameters (e.g., those for image CNN, matching CNN, MLP, $\mathbf{w}_{im}$ and $b_{im}$ in Eq. (\ref{eq_image_cnn}), and word representations) can be jointly learned under the supervision from matching instances. Threefold benefits are provided. Firstly, the image CNN can be tuned to generate a better image representation for matching. Secondly, word representations can be tuned for further composition and matching processes.  Thirdly, the matching CNN (as detailed in the following) composes word representations to different fragments and lets the image representation meet these fragments at different levels, which can fully exploit the inter-modal matching correspondences between image and sentence. With the nonlinear projection in Eq. (\ref{eq_image_cnn}), the image representations $\nu_{im}$ for different matching CNNs are expected to encode the image content for matching the composed semantic fragments of the sentence.

\subsection{Different Variants of Matching CNN}
\label{sec:matchingCNN}

To fully exploit the matching relations of image and sentence, we let the image representation meet and interact with different composed fragments of the sentence (roughly the word, phrase, and sentence) to generate the joint representation.

\subsubsection{Word-level Matching CNN}

\begin{figure}[t!]
\begin{center}
   \includegraphics[width=1\linewidth]{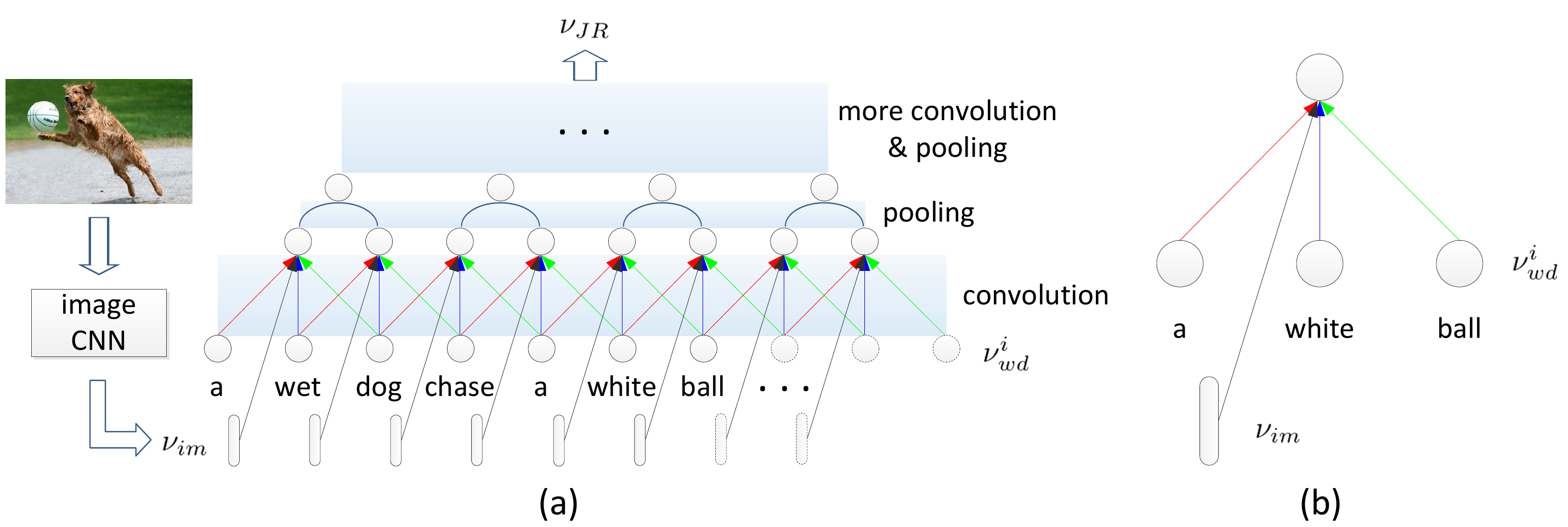}

\end{center}
   \caption{The word-level matching CNN. (a) The word-level matching CNN architecture. (b) The convolution units of multimodal convolution layer of MatchCNN${_{wd}}$.   The dashed ones indicate the zero padded word and image representations, which are gated out after convolution process.}
\label{fig:word}
\end{figure}

In order to find the word-level matching relation, we let the image meet with the word-level fragments of sentence and learn their interactions and relations. Moreover, as most convolutional models \cite{ cnn_speech,cnn_image}, we consider the convolution units with a local ``receptive field" and shared weights to adequately model the rich structures for word composition and inter-modal interaction. The word-level matching CNN, denoted as MatchCNN$_{wd}$, is designed as in Figure \ref{fig:word} (a). After sequential layers of convolution and pooling, the joint representation of image and sentence is generated as the input of MLP for calculating the matching score.

\vspace{-10pt}
\paragraph{Convolution} Generally, with a sequential input $\nu$, the convolution unit for feature map of type-$f$ (among $F_\ell$ of them)  on the $\ell^{th}$ layer is
\begin{equation}
\label{eq_convolution}
\nu_{(\ell, f)}^{i} \overset{\text{def}}{=}
\sigma(\mathbf{w}_{(\ell,f)} \vec{\nu}_{(\ell-1)}^{i} + b_{(\ell,f)}),
\end{equation}
where $\mathbf{w}_{(\ell, f)}$ are the parameters for the $f$ feature map on $\ell^{th}$ layer, $\sigma(\cdot)$ is the activation function, and $\vec{\nu}_{(\ell-1)}^{i}$ denotes the segment of $(\ell\hspace{-1pt}-\hspace{-1pt}1)^{th}$ layer for the convolution at location $i$ , which is defined as follows
\begin{equation}
\label{eq_segment}
\vec{\nu}_{(\ell-1)}^{i}  \overset{\text{def}}{=}  \nu_{(\ell-1)}^{i} \parallel \nu_{(\ell-1)}^{i+1} \parallel \cdots \parallel\nu_{(\ell-1)}^{i+k_{rp}-1}.
\end{equation}
$k_{rp}$ defines the size of local ``receptive field'' for convolution.  ``$\parallel$'' concatenates the neighboring $k_{rp}$ word vectors into a long vector. In this paper, $k_{rp}$ is chosen as 3 for the convolution process.

As MatchCNN$_{wd}$ targets at exploring word-level matching relation, the multimodal convolution layer is introduced by letting the image meet the word-level fragments of sentence. The convolution unit of the multimodal convolution layer is illustrated in Figure \ref{fig:word} (b). The input of the multimodal convolution unit is denoted as:
\begin{equation}
\label{eq_multimodal_word}
\vec{\nu}_{(0)}^{i}  \overset{\text{def}}{=}  \nu_{wd}^{i} \parallel \nu_{wd}^{i+1} \parallel \cdots \parallel\nu_{wd}^{i+k_{rp}-1}\parallel\nu_{im},
\end{equation}
where $\nu_{wd}^{i}$ is the vector representation of word $i$ of the sentence, and $\nu_{im}$ is the encoded image feature for matching word-level fragments of sentence.  It is not hard to see that this input will lead the ``interaction" between words and image representation at the first convolution layer, which provides the local matching signal at word level. From the sentence perspective, the multimodal convolution on $\vec{\nu}_{(0)}^{i}$ composes the words $\nu_{wd}^{i},\cdots, \nu_{wd}^{i+k_{rp}-1}$ in local ``receptive field'' to a higher semantic representation, such as the phrase ``\texttt{\small a white ball}". From the matching perspective, the multimodal convolution on $\vec{\nu}_{(0)}^{i}$ captures and learns the inter-modal correspondence between image representation and the word-level fragments of sentence. The meanings of the word ``\texttt{\small ball}'' and the composed phrase ``\texttt{\small a white ball}" are grounded in the image to make the inter-modal matching relations.

Moreover, in order to handle natural sentences of variable lengthes, the maximum length of sentence is fixed for MatchCNN$_{wd}$. Zero vectors are padded for the image and word representation, as the dashed ones in Figure \ref{fig:word} (a). The output of the convolution process on zero vectors is gated to be zero. The convolution process in Eq. (\ref{eq_convolution}) is further formulated as:
 \begin{equation}
\label{eq_gate}
\begin{split}
\nu_{(\ell, f)}^{i}=g(\vec{\nu}_{(\ell-1)}^{i})\cdot \sigma(\mathbf{w}_{(\ell,f)} \vec{\nu}_{(\ell-1)}^{i} + b_{(\ell,f)})\\
\hbox{where,}  \quad g(x)=
\begin{cases}
0, & x==\textbf{0}\cr
1,& \hbox{otherwise}
\end{cases}
\end{split}
\end{equation}
The gating function can eliminate the unexpected matching noise composed from the convolution process.

\vspace{-10pt}
\paragraph{Max-pooling} After each convolution layer, a max-pooling layer is followed. Taking a two-unit window max-pooling as an example, the pooled feature is obtained by:
 \begin{equation}
\label{eq_maxpooling}
\begin{split}
\nu^i_{(\ell+1,f)} = \max(\nu^{2i}_{(\ell,f)},\nu^{2i+1}_{(\ell,f)})
\end{split}
\end{equation}
The effects of max-pooling are two-fold. 1) Together with the stride as two, the max-pooling process lowers the  dimensionality of the representation by half, thus quickly making the final joint representation of the image and sentence. 2) It helps filter out the undesired interaction and relation between image and fragments of sentence. Take the sentence in Figure \ref{fig:word} (a) as an example, the composed phrase ``\texttt{\small dog chase a}'' matches more closely to the image than ``\texttt{\small chase a white}''. Therefore, we can imagine that a well-trained multimodal convolution unit will generate better matching representation of ``\texttt{\small dog chase a}'' and image. The max-pooling process will pool the matching representation out for further convolution and pooling processes.

The convolution and pooling processes explore and summarize the local matching signals explored at the word level. More layers of convolution and pooling can be further employed to form matching decisions at larger scales and finally reach a global joint representation. Specifically, in this paper another two more convolution and max-pooling layers alternate to summarize the local matching decisions and finally produce the global joint representation of matching, which reflects the inter-modal correspondence between image and word-level fragments of the sentence.

\subsubsection{Phrase-level Matching CNN}
\begin{figure}[t!]
\begin{center}
\includegraphics[width=1\linewidth]{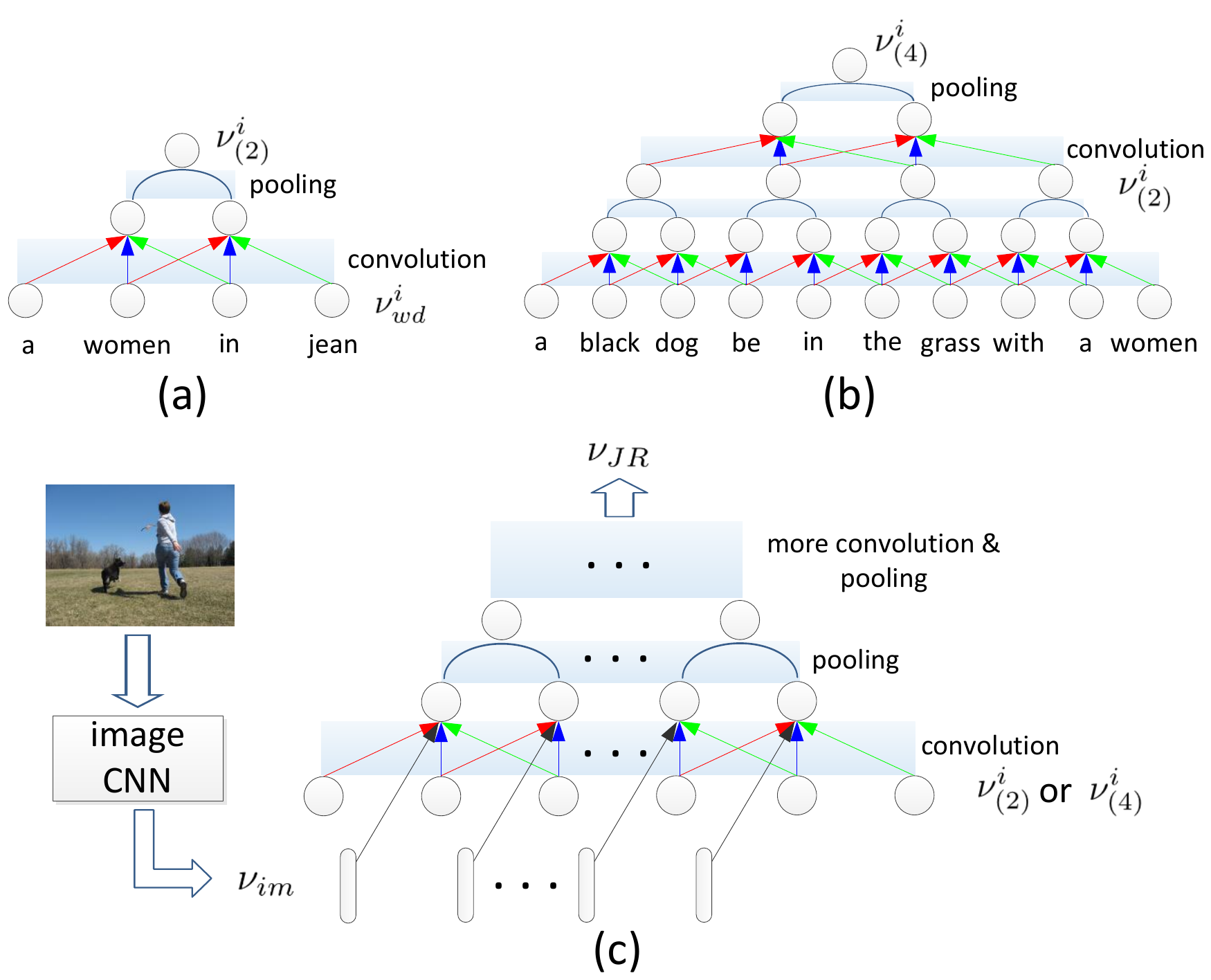}

\end{center}
   \caption{The phrase-level matching CNN and composed phrases. (a): The short phrase is composed by one layer convolution and pooling. (b): The long phrase is composed by two sequential layers of convolution and pooling. (c): The phase-level matching CNN architecture.}
\label{fig:phrase}

\end{figure}

Different from matching CNN at word-level, we let CNN work solely on words to certain levels before interacting with the image.  Without seeing the image feature, the convolution process will compose the words in the ``receptive field'' into a higher semantic representation, while the max-pooling process will filter out the undesired compositions. These composed representations are named as phrase from the language perspective. We let image meet the composed phrases to reason their inter-modal matching relations.

As  illustrated in Figure \ref{fig:phrase} (a), after one layer of convolution and max-pooling process, short phrases (denoted as $\nu_{(2)}^i$) are composed from four words, such as ``\texttt{\small a woman in jean}''. These composed short phrases present richer and specific descriptions about the objects and their relationships compared with single words, such as ``\texttt{\small woman}'' and ``\texttt{\small jean}''. With an additional layer of convolution and max-pooling process on short phrases, long phrases (denoted as $\nu_{(4)}^i$) are composed from four short phrases (also from ten words), such as ``\texttt{\small a black dog be in the grass with a woman}'' in Figure \ref{fig:phrase} (b). Compared with the composed short phrases and single words, the long phrases present even richer and higher semantic meanings about the specific description of the objects, their activities, and their relative positions.

In order to reason the inter-modal relations between image and the composed phrases, a multimodal convolution layer is introduced by performing convolution on the image and phrase representations. The input of the multimodal convolution unit is:
\begin{equation}
\label{eq_multimodal_phrase1}
\vec{\nu}_{ph}^{i}  \overset{\text{def}}{=}  \nu_{ph}^{i} \parallel \nu_{ph}^{i+1} \parallel \cdots \parallel\nu_{ph}^{i+k_{rp}-1}\parallel\nu_{im}.
\end{equation}
where $\nu_{ph}^{i}$ is the composed phrase representation, which can be either short phrases $\nu_{(2)}^i$ or long phrases $\nu_{(4)}^i$. The multimodal convolution process produces the phrase-level matching decisions. Then the layers after that (namely the max-pooling layer or convolution layer) can be viewed as further fusion of these local phrase-level matching decisions to a  joint representation, which captures the local matching relations between image and composed phrase fragments. Specifically, for short phrases, two sequential layers of convolution and pooling are followed to generate the  joint representation. We name the matching CNN for short phrases and image as MatchCNN$_{phs}$. For long phrases, only one sequential layer of convolution and pooling is used to summarize the local matching to the joint representation. The matching CNN for long phrases and image is named as MatchCNN$_{phl}$.

\subsubsection{Sentence-level Matching CNN}
\label{sec_sentence}

\begin{figure}[t]
\begin{center}
   \includegraphics[width=1\linewidth]{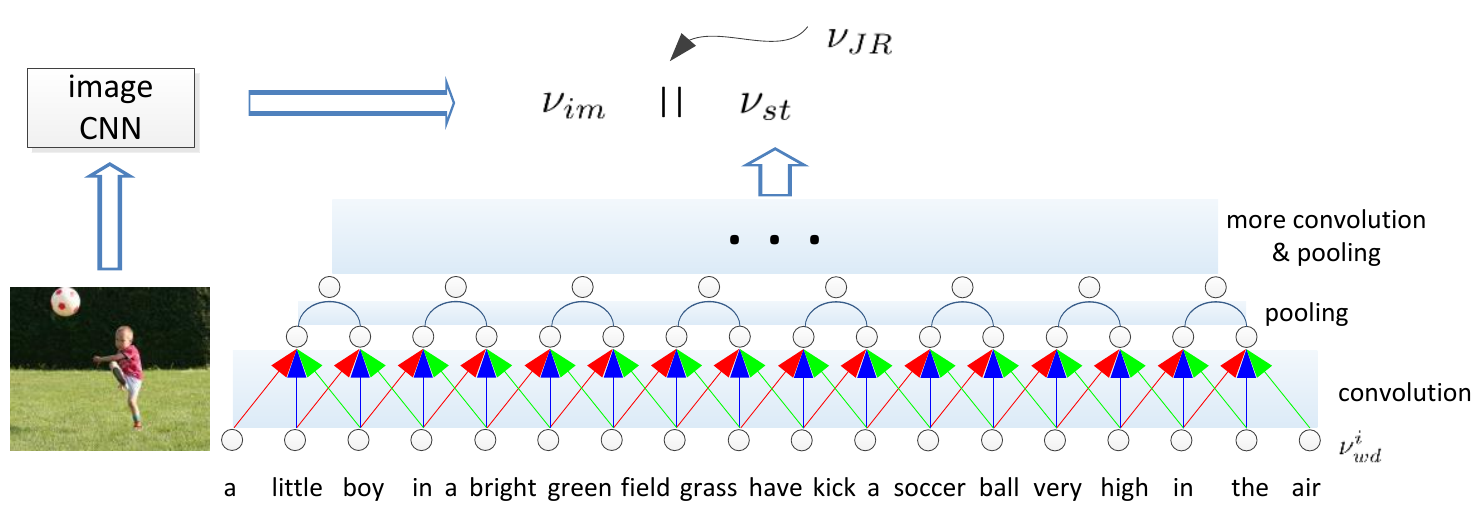}

\end{center}
   \caption{The sentence-level matching CNN. The joint representation is obtained by concatenating the image and sentence representations together.}
\label{fig:sentence}
\end{figure}

The sentence-level convolutional  matching CNN, denoted as MatchCNN$_{st}$, goes one step further in the composition and defers the matching  until the sentence is fully represented, as illustrated in Figure \ref{fig:sentence}. More specifically, one image CNN encodes the image into a feature vector. One sentence CNN, consisting of three sequential layers of convolution and pooling, represents the whole sentence as a feature vector. The multimodal layer concatenates the image and sentence representation together as their joint representation:
\begin{equation}
\label{eq_word_image_concatenation}
\begin{split}
{\nu_{JR}} = \nu_{im} \parallel \nu_{st},
\end{split}
\end{equation}
where $\nu_{st}$ denotes the sentence representation by vectorizing the features in the last layer of the sentence CNN.

For the sentence ``\texttt{\small a little boy in a bright green field have kick a soccer ball very high in the air}" illustrated in Figure \ref{fig:sentence}, although word-level and phrase-level fragments, such as ``\texttt{\small boy}", ``\texttt{\small kick a soccer ball}", correspond to the objects  as well as  their activities in the image, the whole sentence needs to be fully represented to make a reliable association with the image. The sentence CNN with layers of convolution and pooling is used to encode the whole sentence as a feature vector representing its semantic meaning. Concatenating the image and sentence representation together,  MatchCNN$_{st}$ does no non-trivial matching, but transfer the representations of the two modalities to the later MLP for fusing and matching.

\subsection{$m$-CNNs with Different Matching CNNs}

We can get different $m$-CNNs with different variants of Matching CNNs, namely $m$-CNN$_{wd}$, $m$-CNN$_{phs}$, $m$-CNN$_{phl}$, and $m$-CNN$_{st}$. To fully exploit the inter-modal matching relations between image and sentence at different levels, we use an ensemble $m$-CNN$_{ENS}$ of the four variants  by summing the matching scores generated from these $m$-CNNs together.

\begin{table}\footnotesize
\begin{center}
\begin{tabular}{|c|c|c|c|}
\hline
MatchCNN$_{wd}$&MatchCNN$_{phs}$&MatchCNN$_{phl}$&MatchCNN$_{st}$\\
\hline\hline
\textbf{+ $\nu_{im}$}& & & \\
\hline\hline
multi-conv-200&conv-200&conv-200&conv-200\\
\hline\hline
max-2 &max-2 &max-2&max-2 \\
\hline\hline
& \textbf{+ $\nu_{im}$}&& \\
\hline\hline
conv-300 &multi-conv-300 &conv-300&conv-300 \\
\hline\hline
max-2 &max-2&max-2&max-2 \\
\hline\hline
& & \textbf{+ $\nu_{im}$}& \\
\hline\hline
conv-300&conv-300&multi-conv-300&conv-300 \\
\hline\hline
max-2&max-2&max-2&max-2 \\
\hline\hline
&&&\textbf{+ $\nu_{im}$} \\
\hline
\end{tabular}
\end{center}
\caption{Configurations of MatchCNN$_{wd}$, MatchCNN$_{phs}$, MatchCNN$_{phl}$, and MatchCNN$_{st}$ in columns. (conv denotes convolution layer; multi-conv denotes the multimodal convolution layer; max denotes max pooling layer.)}
\label{table:configuration}
\end{table}

\section{Implementation details}
In this section, we describe the detailed configurations of our proposed $m$-CNN models and how we train the proposed networks.
\subsection{Configurations}


We use two different image CNNs,  OverFeat \cite{sermanet_arxiv2014} (the ``fast'' network) and VGG \cite{simonyan_arxiv2014} (with 19 weight layers), with which we take not only the architecture but also the original parameters (learnt on ImageNet dataset) for initialization. By chopping the top softmax layer and the last ReLU layer, the output of the last fully-connected layer is deemed as image representation, denoted as $CNN_{im}(I)$  in Eq. (\ref{eq_image_cnn}).

The configurations of MatchCNN$_{wd}$, MatchCNN$_{phs}$, MatchCNN$_{phl}$, and MatchCNN$_{st}$ are outlined in Table \ref{table:configuration}. We use three convolution layers, three max pooling layers, and an MLP with two fully connected layers for all these four networks. The first convolution layer of MatchCNN$_{wd}$, second convolution layer of MatchCNN$_{phs}$, and third convolution layer of MatchCNN$_{phl}$ are the multimodal convolution layers, which blend the image representation and fragments of the sentence together to compose a higher level semantic representation. The MatchCNN$_{st}$ concatenates the image and sentence representation together and leave the interaction to the final MLP. The matching CNNs are designed on fixed architectures, which need to be set to accommodate the maximum length of the input sentences. During our evaluations, the maximum length is set as 30. The word representations are initialized by the skip-gram model \cite{mikolov_iclr2013} with dimension 50. The joint representation obtained from the matching CNNs is fed into MLP with one hidden layer with size 400.



\subsection{Learning}
\label{sec_learning}
The $m$-CNN models can be trained with contrastive sampling using a ranking loss function. More specifically, for the score function $s_{match}(\cdot)$ as in Eq. (\ref{eq_mlp}), the objective function is defined as:

\begin{equation}
\label{eq_objective}
\begin{split}
\small
e_{\theta}(x_n,y_n,y_m) = \hspace{-4pt} \qquad \qquad \quad \qquad \qquad \qquad \qquad   \\
\max \big(0,\mu\hspace{-3pt}- \hspace{-3pt}s_{match}(x_n,y_n)\hspace{-3pt}+\hspace{-3pt}s_{match}(x_n,y_m)\big)
\end{split}
\end{equation}

where $\theta$ denotes the parameters, $(x_n,y_n)$ denotes the correlated image-sentence pair, and $(x_n,y_m)$ is the randomly sampled uncorrelated image-sentence pair ($n\neq m$). The notational meaning of $x$ and $y$ varies with the matching task: for image retrieval from query sentence, $x$ denotes the natural sentence and $y$ denotes the image; for sentence retrieval from query image, it is just the opposite. The object is to force the matching score of the correlated pair $(x_n,y_n)$ to be greater than the uncorrelated pair $(x_n,y_m)$ by a margin $\mu$, which is simply set as 0.5 for our training process.


We use stochastic gradient descent (SGD) with mini-batches of 100$\sim$150 for optimization.
In order to avoid overfitting, early-stopping \cite{caruana_nips2000} and dropout (with probability 0.1) \cite{hinton_corr2012} are used. ReLU is used as the activation function throughout $m$-CNNs.

\begin{table*} \small
\begin{center}
    \begin{tabular}{c|cccc|cccc}
    \hline
          & \multicolumn{4}{|c|}{Sentence Retrieval} & \multicolumn{4}{|c}{Image Retrieval}  \\
          \hline
          \hline
        & R@1 & R@5 & R@10 & Med $r$ & R@1 & R@5 & R@10 & Med $r$ \\
        \hline
        \hline
        Random Ranking & 0.1 & 0.6 & 1.1 & 631 & 0.1 & 0.5  & 1.0 & 500\\
        DeViSE \cite{frome_nips2013}& 4.8 & 16.5 & 27.3 & 28.0 & 5.9 & 20.1  & 29.6 & 29\\
        SDT-RNN \cite{socher_tacl2014}& 6.0 & 22.7 & 34.0 & 23.0 & 6.6 & 21.6  & 31.7 & 25\\
        MNLM \cite{kiros_2014}& 13.5 & 36.2 & 45.7 & 13 & 10.4 & 31.0  & 43.7 & 14\\
        MNLM-vgg \cite{kiros_2014}& 18.0 & 40.9 & 55.0 & 8 & 12.5 & 37.0  & 51.5 & 10\\
        $m$-RNN \cite{mao_2014}& 14.5 & 37.2 & 48.5 & 11 & 11.5 & 31.0  & 42.4 & 15\\
        Deep Fragment  \cite{karpathy_2014} & 12.6 & 32.9 & 44.0 &14 & 9.7 & 29.6  & 42.5 & 15\\
        RVP (T) \cite{chen_2014}& 11.6 & 33.8 & 47.3 & 11.5 & 11.4 & 31.8  & 45.8 & 12.5\\
        RVP (T+I) \cite{chen_2014}& 11.7 & 34.8 & 48.6 & 11.2 & 11.4 & 32.0  & 46.2 & 11\\
        DVSA (DepTree) \cite{karpathy_dvsa_2014}& 14.8 & 37.9 & 50.0 & 9.4 & 11.6 & 31.4  & 43.8 & 13.2\\
        DVSA (BRNN) \cite{karpathy_dvsa_2014}& 16.5 & 40.6 & 54.2 & 7.6 & 11.8 & 32.1  & 44.7 & 12.4\\
        DCCA \cite{yan_cvpr2015} & 17.9 & 40.3 & 51.9 & 9 & 12.7 & 31.2 & 44.1 & 13\\
        NIC \cite{vinyals_2014}& {20.0} & * & {61.0} & {6} & {19.0} & *  & {64.0} & \textbf{5}\\

        FV (Mean Vec) \cite{klein_cvpr2015} & 22.6 & 48.8 & 61.2 & 6 & 19.1 & 45.3 & 60.4 & 7 \\
        FV (GMM) \cite{klein_cvpr2015} & 28.4 & 57.7 & 70.1 & 4 & 20.6 & 48.5 & 64.1 & 6 \\
        FV (LMM) \cite{klein_cvpr2015} & 27.7 & 56.6 & 69.0 & 4 & 19.8 & 47.6 & 62.7 & 6 \\
        FV (HGLMM) \cite{klein_cvpr2015} & 28.5 & 58.4 & 71.7 & 4 & 20.6 & 49.4 & 64 & 6 \\
        FV (GMM+HGLMM) \cite{klein_cvpr2015} & \textbf{31.0} & \textbf{59.3} & \textbf{73.7} & \textbf{4} & \textbf{21.3} & \textbf{50.0} & \textbf{64.8} & \textbf{5} \\

        \hline
        \hline
        OverFeat \cite{sermanet_arxiv2014}:&&&&&&&& \\
        $m$-CNN$_{wd}$ & 8.6 &26.8 &38.8&18.5 &8.1 &24.7 &36.1&20\\
        $m$-CNN$_{phs}$ & 10.5 &29.4 &41.7&15 &9.3 &27.9 &39.6&17\\
        $m$-CNN$_{phl}$ & 10.7 &26.5 &38.7&18 &8.1 &26.6 &37.8&18\\
        $m$-CNN$_{st}$ & 10.6 &32.5 &43.6&14 &8.5 &27.0 &39.1&18\\
        $m$-CNN$_{ENS}$ & {14.9} &{35.9} &{49.0}&{11.0} &{11.8} &{34.5} &{48.0}&{11.0}\\
        \hline
        \hline
        VGG \cite{simonyan_arxiv2014}:&&&&&&&& \\
        $m$-CNN$_{wd}$ & 15.6 &40.1 &55.7&8 &14.5 & 38.2 &52.6&9\\
        $m$-CNN$_{phs}$ & 18.0 & 43.5 &57.2&8 &14.6 &39.5 &53.8&9\\
        $m$-CNN$_{phl}$ & 16.7 & 43.0 &56.7&7 &14.4 &38.6 &52.2&9\\
        $m$-CNN$_{st}$ & 18.1 &44.1 &57.9&7 &14.6 &38.5 &53.5&9\\
        $m$-CNN$_{ENS}$ &  {24.8} & {53.7}& {67.1}& {5} & {20.3} & {47.6} & {61.7}& \textbf{5}\\
        \hline
        \hline
    \end{tabular}

\end{center}
\caption{Bidirectional image and sentence retrieval results on Flickr8K.}
\label{table:flickr8k}
\end{table*}

\begin{table*} \small
\begin{center}
    \begin{tabular}{c|cccc|cccc}
    \hline
          & \multicolumn{4}{|c|}{Sentence Retrieval} & \multicolumn{4}{|c}{Image Retrieval}  \\
          \hline
          \hline
        & R@1 & R@5 & R@10 & Med $r$ & R@1 & R@5 & R@10 & Med $r$ \\
        \hline
        \hline
        Random Ranking & 0.1 & 0.6 & 1.1 & 631 & 0.1 & 0.5  & 1.0 & 500\\
        DeViSE \cite{frome_nips2013}& 4.5 & 18.1 & 29.2 & 26 & 6.7 & 21.9  & 32.7 & 25\\
        SDT-RNN \cite{socher_tacl2014}& 9.6 & 29.8 & 41.1 & 16 & 8.9 & 29.8  & 41.1 & 16\\
        MNLM \cite{kiros_2014}& 14.8 & 39.2 & 50.9 & 10 & 11.8 & 34.0  & 46.3 & 13\\
        MNLM-vgg \cite{kiros_2014}& 23.0 & 50.7 & 62.9 & 5 & 16.8 & 42.0  & 56.5 & 8\\
        $m$-RNN \cite{mao_2014}& 18.4 & 40.2 & 50.9 & 10 & 12.6 & 31.2  & 41.5 & 16\\
        $m$-RNN-vgg \cite{mao_iclr_2015}& {35.4} & {63.8} & {73.7} & \textbf{3} & {22.8} & {50.7}  & {63.1} & {5}\\
        Deep Fragment  \cite{karpathy_2014}& 14.2 & 37.7 & 51.3 &10 & 10.2 & 30.8  & 44.2 & 14\\
        RVP (T) \cite{chen_2014}& 11.9 & 25.0 & 47.7 & 12 & 12.8 & 32.9  & 44.5 & 13\\
        RVP (T+I) \cite{chen_2014}& 12.1 & 27.8 & 47.8 & 11 & 12.7 & 33.1  & 44.9 & 12.5\\
        DVSA (DepTree)  \cite{karpathy_dvsa_2014}& 20.0 & 46.6 & 59.4 & 5.4 & 15.0 & 36.5  & 48.2 & 10.4\\
        DVSA (BRNN)  \cite{karpathy_dvsa_2014}& 22.2 & 48.2 & 61.4 & 4.8 & 15.2 & 37.7  & 50.5 & 9.2\\
        DCCA \cite{yan_cvpr2015} & 16.7 & 39.3 & 52.9 & 8 & 12.6 & 31.0 & 43.0 & 15 \\
        NIC \cite{vinyals_2014}& 17.0 & * & 56.0 & 7 & 17.0 & *  & 57.0 & 7\\
        LRCN \cite{donahue_2014}& * & * & * & * & 17.5 & 40.3  & 50.8 & 9\\

        RTP (joint training) \cite{plummer_2015} & 31.0 & 58.6 & 67.9 & * & 22.0 & 50.7 & 62.0 & * \\
        RTP (SAE) \cite{plummer_2015} & 36.7 & 61.9 & 73.6 & * & 25.4 & 55.2 & 68.6 & *\\
        RTP (weighted distance) \cite{plummer_2015} & \textbf{37.4} & 63.1 & 74.3 & * & 26.0 & 56.0 & 69.3 & * \\

        FV (Mean Vec) \cite{klein_cvpr2015} & 24.8 & 52.5 & 64.3 & 5 & 20.5 & 46.3 & 59.3 & 6.8 \\
        FV (GMM) \cite{klein_cvpr2015} & 33.0 & 60.7 & 71.9 & \textbf{3} & 23.9 & 51.6 & 64.9 & 5 \\
        FV (LMM) \cite{klein_cvpr2015} & 32.5 & 59.9 & 71.5 & 3.2 & 23.6 & 51.2 & 64.4 & 5 \\
        FV (HGLMM) \cite{klein_cvpr2015} & 34.4 & 61.0 & 72.3 & \textbf{3} & 24.4 & 52.1 & 65.6 & 5 \\
        FV (GMM+HGLMM) \cite{klein_cvpr2015} & 35.0 & 62.0 & 73.8 & \textbf{3} & 25.0 & 52.7 & 66.0 & 5 \\

        \hline
        \hline
        OverFeat \cite{sermanet_arxiv2014}:&&&&&&&& \\
        $m$-CNN$_{wd}$ & 12.7 & 30.2 & 44.5 & 14 & 11.6 & 32.1 & 44.2 & 14\\
        $m$-CNN$_{phs}$& 14.4 & 38.6 & 49.6 & 11 & 12.4 & 33.3 & 44.7 & 14\\
        $m$-CNN$_{phl}$& 13.8 & 38.1 & 48.5 & 11.5&11.6 & 32.7 & 44.1 & 14\\
        $m$-CNN$_{st}$ & 14.8 & 37.9 & 49.8 & 11 & 12.5 & 32.8 & 44.2 & 14\\
        $m$-CNN$_{ENS}$ & 20.1 & 44.2 & 56.3 & 8 & 15.9 & 40.3 & 51.9 & 9.5\\
        \hline
        \hline
        VGG \cite{simonyan_arxiv2014}:&&&&&&&& \\
        $m$-CNN$_{wd}$ & 21.3 & 53.2 & 66.1 & 5 & 18.2 & 47.2 & 60.9 & 6\\
        $m$-CNN$_{phs}$& 25.0 & 54.8 & 66.8 & 4.5 & 19.7 & 48.2 & 62.2 & 6\\
        $m$-CNN$_{phl}$& 23.9 & 54.2 & 66.0 & 5 & 19.4 & 49.3 & 62.4 & 6\\
        $m$-CNN$_{st}$ & 27.0 & 56.4 & 70.1 & 4 & 19.7 & 48.4 & 62.3 & 6\\
        $m$-CNN$_{ENS}$ & {33.6} & \textbf{64.1} & \textbf{74.9} & \textbf{3} & \textbf{26.2} & \textbf{56.3} & \textbf{69.6} & \textbf{4}\\
        \hline
        \hline

    \end{tabular}

\end{center}
\caption{Bidirectional image and sentence retrieval results on Flickr30K.}
\label{table:flickr30k}
\end{table*}

\begin{table*} \small
\begin{center}
    \begin{tabular}{c|cccc|cccc}
    \hline
          & \multicolumn{4}{|c|}{Sentence Retrieval} & \multicolumn{4}{|c}{Image Retrieval}  \\
          \hline
          \hline
        & R@1 & R@5 & R@10 & Med $r$ & R@1 & R@5 & R@10 & Med $r$ \\
        \hline
        \hline
        Random Ranking & 0.1 & 0.6 & 1.1 & 631 & 0.1 & 0.5  & 1.0 & 500\\
        $m$-RNN-vgg \cite{mao_iclr_2015}& 41.0 & 73.0 & 83.5 & 2 & 29.0 & 42.2  & 77.0 & 3\\
        DVSA\cite{karpathy_dvsa_2014}& 38.4 & 69.9 & 80.5 & \textbf{1} & 27.4 & 60.2  & 74.8 & 3\\

        STV (uni-skip) \cite{kiros_2015} & 30.6 & 64.5 & 79.8 & 3 & 22.7 & 56.4 & 71.7 & 4 \\
        STV (bi-skip) \cite{kiros_2015} & 32.7 & 67.3 & 79.6 & 3 & 24.2 & 57.1 & 73.2 & 4 \\
        STV (combine-skip) \cite{kiros_2015} & 33.8 & 67.7 & 82.1 & 3 & 25.9 & 60.0 & 74.6 & 4 \\

        FV (Mean Vec) \cite{klein_cvpr2015} & 33.2 & 61.8 & 75.1 & 3 & 24.2 & 56.4 & 72.4 & 4 \\
        FV (GMM) \cite{klein_cvpr2015} & 39.0 & 67.0 & 80.3 & 3& 24.2 & 59.2 & 76.0 & 4 \\
        FV (LMM) \cite{klein_cvpr2015} & 38.6 & 67.8 & 79.8 & 3 & 25.0 & 59.5 & 76.1 & 4 \\
        FV (HGLMM) \cite{klein_cvpr2015} & 37.7 & 66.6 & 79.1 & 3 & 24.9 & 58.8 & 76.5 & 4 \\
        FV (GMM+HGLMM) \cite{klein_cvpr2015} & 39.4 & 67.9 & 80.9 & 2 & 25.1 & 59.8 & 76.6 & 4 \\

        \hline
        \hline
        VGG \cite{simonyan_arxiv2014}:&&&&&&&& \\
        $m$-CNN$_{wd}$ & 34.1 & 66.9 & 79.7 & 3 & 27.9 & 64.7 & 80.4 & 3\\
        $m$-CNN$_{phs}$& 34.6 & 67.5 & 81.4 & 3 & 27.6 & 64.4 & 79.5 & 3\\
        $m$-CNN$_{phl}$& 35.1 & 67.3 & 81.6 & 2 & 27.1 & 62.8 & 79.3 & 3\\
        $m$-CNN$_{st}$ & 38.3 & 69.6 & 81.0 & 2 & 27.4 & 63.4 & 79.5 & 3\\
        $m$-CNN$_{ENS}$ & \textbf{42.8} & \textbf{73.1} & \textbf{84.1} & {2} & \textbf{32.6} & \textbf{68.6} & \textbf{82.8} & \textbf{3}\\
        \hline
        \hline

    \end{tabular}

\end{center}
\caption{Bidirectional image and sentence retrieval results on Microsoft COCO.}
\label{table:mscoco}
\end{table*}

\section{Experiments}
\label{sec_experiment}

In this section, we evaluate the effectiveness of our $m$-CNNs on bidirectional image and sentence retrieval. We begin by describing the datasets used for evaluation, followed by a brief description of competitor models. As our $m$-CNNs are bidirectional, we evaluate the performances on both  image retrieval and sentence retrieval.

\subsection{Datasets}

\label{sec_setting:database}
We test our matching models on the public image-sentence datasets, with varying sizes and characteristics.

\vspace{4pt}
\noindent \textbf{Flickr8K} \cite{hodosh_jair2013}  This dataset consists of 8,000 images collected from Flickr. Each image is accompanied with 5 sentences describing the image content. This database provides the standard training, validation, and testing split.

\vspace{4pt}
\noindent \textbf{Flickr30K} \cite{young_tacl2014} This dataset  consists of 31,783 images collected from Flickr. Each image is also accompanied with 5 sentences describing the content of the image. Most of the images depict varying human activities. We used the public split as in \cite{mao_2014} for training, validation, and testing.

\vspace{4pt}
\noindent \textbf{Microsoft COCO} \cite{lin_arxiv2014} This dataset consists of 82,783 training and 40,504 validation images with 80 categories labeled for a total of 886,284 instances. Each image is also associated with 5 sentences describing the content of the image. We used the public split as in \cite{mao_iclr_2015} for training, validation, and testing.

\subsection{Competitor Models}
\label{sec_competitors}


We {compared} our models with recently developed models on the performances of the bidirectional image and sentence retrieval, specifically DeViSE  \cite{frome_nips2013}, SDT-RNN \cite{socher_tacl2014}, DCCA \cite{yan_cvpr2015}, FV \cite{klein_cvpr2015}, STV \cite{kiros_2015}, RTP \cite{plummer_2015}, Deep Fragment \cite{karpathy_2014}, $m$-RNN \cite{mao_iclr_2015,mao_2014}, MNLM  \cite{kiros_2014}, RVP \cite{chen_2014}, DVSA \cite{karpathy_dvsa_2014}, NIC \cite{vinyals_2014}, and LRCN \cite{donahue_2014}. DeViSE and Deep Fragment are regarded as working on word-level and phrase-level, respectively. SDT-RNN, DCCA, and FV are all regarded as working on the sentence-level, which embed the image and sentence into the same semantic space. The other models, namely MNLM, $m$-RNN, RVP, DVSA, NIC, and LRCN, which are originally proposed for automatic image captioning,
can also be used for retrieval in both directions.

\begin{table*} \scriptsize
\begin{center}

    \renewcommand{\multirowsetup}{\centering}
    \begin{tabular}{c|c|cccc}
    \hline

     image   &sentence & {$m$-CNN$_{wd}$} & {$m$-CNN$_{phs}$} & {$m$-CNN$_{phl}$} & {$m$-CNN$_{st}$}\\
        \hline
        \hline
  \multirow{8}{*}{\includegraphics[width=0.20\linewidth]{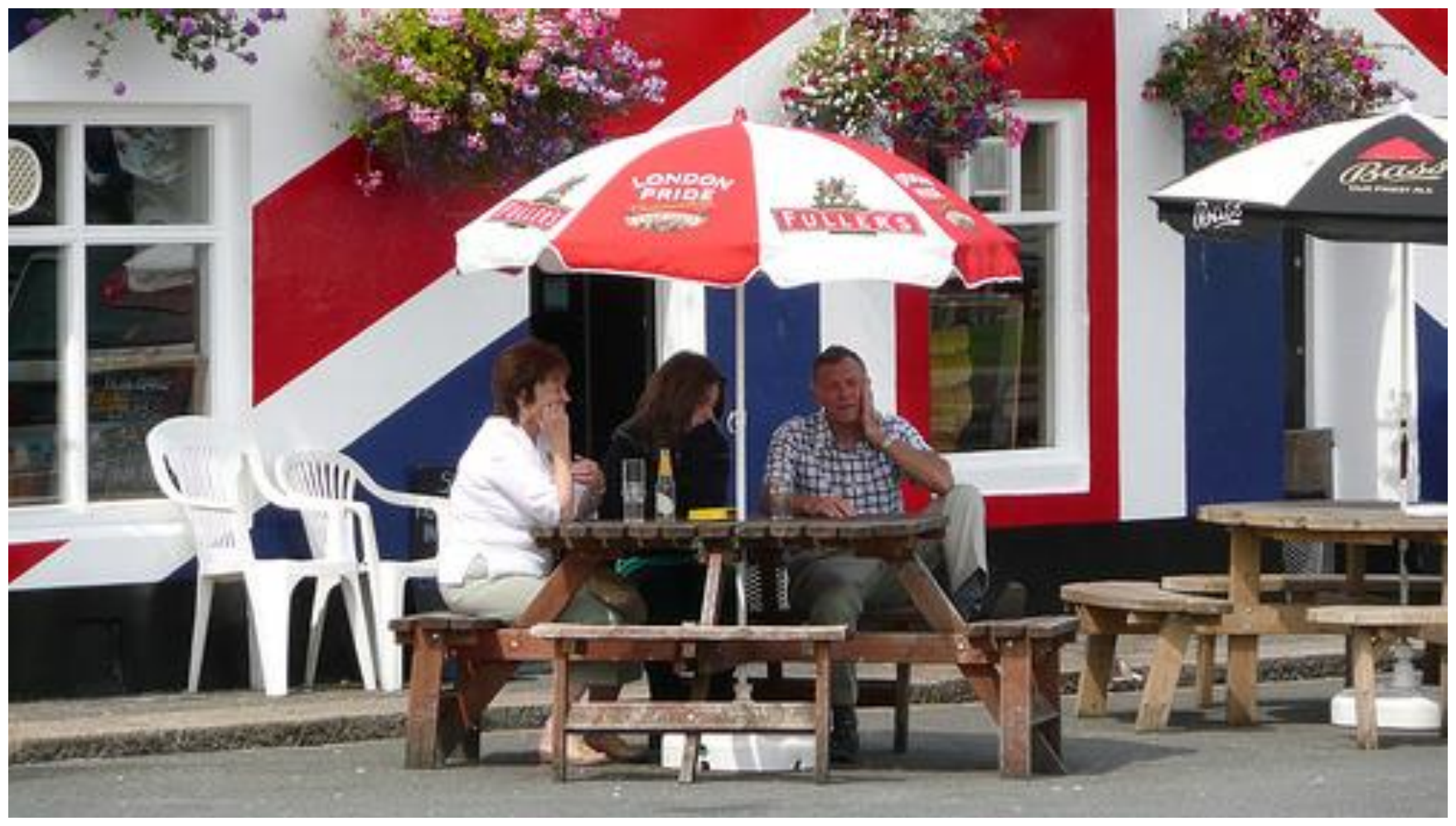}} &\texttt{\textbf{three person sit at an outdoor table in front}} & \multirow{2}{*}{\textbf{-0.87}} & \multirow{2}{*}{\textbf{1.91}} & \multirow{2}{*}{\textbf{-1.84}} & \multirow{2}{*}{\textbf{2.93}}  \\
  &\texttt{\textbf{of a building paint like the union jack .}} & & & &  \\
  \cline{2-6}
  &\texttt{like union at in sit three jack the person a}&\multirow{2}{*}{-1.49} & \multirow{2}{*}{1.66} & \multirow{2}{*}{-3.00} & \multirow{2}{*}{2.37}  \\
  &\texttt{paint building table outdoor of front an .} & & & &  \\
  \cline{2-6}
  &\texttt{sit union a jack three like in of paint the}&\multirow{2}{*}{-2.44} & \multirow{2}{*}{1.55} & \multirow{2}{*}{-3.90} & \multirow{2}{*}{2.53}  \\
  &\texttt{person table outdoor building front at an .} & & & &  \\
 \cline{2-6}
  &\texttt{table sit three paint at a building of like }&\multirow{2}{*}{-1.93} & \multirow{2}{*}{1.64} & \multirow{2}{*}{-3.81} & \multirow{2}{*}{2.52} \\
  &\texttt{the an person front outdoor jack union in .} & & & &  \\
  \hline
  \hline

    \end{tabular}

\end{center}
\caption{The matching scores of the image and sentence. The natural sentence (in bold) is the true caption of the image, while the other three sentences are generated by random reshuffle of words.}
\label{table:composition}
\vspace{-10pt}
\end{table*}

\subsection{Experimental Results and Analysis}

\label{sec_results}
\subsubsection {Bidirectional Image and Sentence Retrieval}

We adopt the evaluation metrics \cite{karpathy_2014} for a fair comparison. More specifically, for bidirectional retrieval, we report the median rank (Med $r$ ) of the closest ground truth result in the list, as well as the R@$K$ (with $K= 1,5,10$) which computes the fraction of times the correct result was found among the top $K$ items. The performances of the proposed $m$-CNNs on bidirectional image and sentence retrieval of  Flickr8K, Flickr30K, Microsoft COCO are illustrated in Table \ref{table:flickr8k}, \ref{table:flickr30k}, and \ref{table:mscoco}. We highlight the best performance of each evaluation metric.


On Flickr8K, FV performs the best, suggesting the strong and beneficial bias of Fisher vector on modeling sentences, which is most obvious when the training data are relatively scarce. Our proposed $m$-CNN performs inferiorly to FV, but still superior to other methods. The reason, as suggested by the results of larger datasets (Flickr30K and Microsoft COCO), is mainly the insufficient training samples. Flickr8K consists of only 8,000 images, which are insufficient for adequately tuning the parameters of the convolutional architectures in $m$-CNNs. On Flickr30K and Microsoft COCO datasets, with more training samples, $m$-CNN$_{ENS}$ (with VGG) outperforms all the competitor models in terms of most metrics, as illustrated in Table. \ref{table:flickr30k} and \ref{table:mscoco}. Moreover, except FV, only NIC slightly outperforms $m$-CNN$_{ENS}$ (with VGG) on image retrieval task measured by R@10. Except the lack of training samples, another possible reason is that NIC uses a better image CNN \cite{szegedy_arxiv2014}, compared with VGG. As discussed in Section \ref{sec:experiment:imagecnn}, the performance of image CNN greatly affects the performance of the bidirectional image and sentence retrieval.

On Flickr30K, with more training instances (30,000 images), the best performing competitor model becomes the RTP on both tasks. Only $m$-RNN-vgg, FV, and RTP outperform $m$-CNN$_{ENS}$ (with VGG) on sentence retrieval task measured by R@1. When it comes to image retrieval, $m$-CNN$_{ENS}$ (with VGG) is consistently better than all competitor models. One possible reason may be that $m$-RNN-vgg is designed for caption generation and is particularly good at finding the suitable sentence for any given image. One possible reason for RTP may be that the Flickr30K entities are specifically presented, where the bounding boxes corresponding to each entity are manually labeled. As such, much more information are available for image retrieval.

On Microsoft COCO, with more training instances (over 110,000 images), the performances of our proposed $m$-CNN in terms of all the evaluation metrics have been significantly improved, compared with those on Flickr8k and Flickr30K. Firstly, it demonstrates that with sufficient training samples, the parameters of the convolutional architecture in $m$-CNN can be more adequately tuned. Secondly, only DVSA outperforms the proposed $m$-CNN$_{ENS}$ (with VGG) on sentence retrieval in terms of Med $r$. On image retrieval, $m$-CNN$_{ENS}$ significantly and consistently outperforms all the competitor models.

%
%
%
%
%

\vspace{-5pt}
\subsubsection {Performances of Different $m$-CNNs}

The proposed $m$-CNN$_{wd}$ and DeViSE \cite{frome_nips2013} both target at exploiting word-level inter-modal correspondences between image and sentence. However, DeViSE treats each word equally and average their word vectors as the representation of the sentence, while our $m$-CNN$_{wd}$ let image interact with each word and compose them to higher semantic representations, which significantly outperforms DeViSE. On the other end, both SDT-RNN \cite{socher_tacl2014} and the proposed $m$-CNN$_{st}$ exploit the matching between image and sentence at the sentence level. However, SDT-RNN encodes each sentence recursively into a feature vector based on a pre-given dependency tree, while $m$-CNN$_{st}$ works on a more flexible manner with sliding window on the sentence to finally generate the sentence representation. Therefore, a better performance is obtained by  $m$-CNN$_{st}$.

Deep Fragment \cite{karpathy_2014} and the proposed $m$-CNN$_{phs}$ and $m$-CNN$_{phl}$ match the image and sentence fragments at phrase levels. However, Deep Fragment uses edges of dependency tree to model the sentence fragments, making it unable to describe more complex relations in sentence. For example, Deep Fragment parses a relative complex phrase ``\texttt{\small black and brown dog}'' to two relations  ``\texttt{\small (CONJ, black, brown)}" and ``\texttt{\small (AMOD, brown, dog)}", while $m$-CNN$_{phs}$ handles the same phrase as a whole to compose them to a higher semantic representation. Moreover, $m$-CNN$_{phl}$ can readily handle longer phrases and reason their grounding meanings in the image. Consequently, better performances of $m$-CNN$_{phs}$ and $m$-CNN$_{phl}$ (with VGG) are obtained compared with Deep Fragment.

Moreover, it can be observed that $m$-CNN$_{st}$ consistently outperform other $m$-CNNs. The sentence CNN can well summarize the natural sentence and make a better sentence-level association with image in $m$-CNN$_{st}$. Other $m$-CNNs captures the matching relations at word and phrase levels. The matching relations should be considered together to fully depict the inter-modal correspondences between image and sentence. Thus $m$-CNN$_{ENS}$ achieves the best performances, which indicates that $m$-CNNs at different levels are complementary with each other to capture the complicated image and sentence matching relations.

\vspace{-3pt}
\subsubsection {Influence of Image CNN}
\label{sec:experiment:imagecnn}

We use OverFeat and VGG to initialize the image CNN in $m$-CNN for the retrieval tasks. It can be observed that $m$-CNNs with VGG significantly outperform that with OverFeat by a large margin, which is consistent with their performance on classification on ImageNet (14\% and 7\% top-5 classification errors for OverFeat and VGG, respectively). Clearly the retrieval performance depends heavily on the efficacy of the image CNN, which might explain the good performance of NIC on Flickr8K. Moreover, region with CNN features \cite{girshick_cvpr2014} are used for encoding image regions to feature vectors, which are used as the image fragments in Deep Fragment and DVSA. In the future, we will consider to incorporate these image CNNs into our $m$-CNNs to make more accurate inter-modal matching.

\vspace{-2pt}
\subsubsection {Composition Abilities of $m$-CNNs}
\label{sec:experiment:composition}


$m$-CNNs can compose words to different semantic fragments of the sentence for the inter-modal matching at different levels, and therefore posses the ability of word composition. More specifically, we want to check whether the $m$-CNNs can compose words of random orders into semantic fragments for matching the image content. As demonstrated in Table \ref{table:composition}, the matching scores between an image and its accompanied sentence (from different $m$-CNNs) greatly decrease after the random reshuffle of words. It is a fairly strong evidence that $m$-CNNs will compose words in natural sequential order into high semantic representations and thus make the inter-modal matching relations between image and sentence.

\section{Conclusion}

We proposed multimodal convolutional neural networks ($m$-CNNs) for matching image and sentence. The proposed $m$-CNNs rely on convolution architectures to compose different semantic fragments of the sentence and learn the interaction between image and the composed fragments at different levels, therefore fully exploit the inter-modal matching relations. Experimental results on bidirectional image and sentence retrieval demonstrate the consistent state-of-the-art performances of our proposed models.

{\small
\bibliographystyle{ieee}
\bibliography{egbib}
}
\end{document}